\documentclass[lettersize,journal]{IEEEtran}

  \usepackage[noadjust]{cite}

\usepackage{float}
\ifCLASSINFOpdf
  \usepackage[pdftex]{graphicx}
\else
  \usepackage[dvips]{graphicx}
\fi

\usepackage{amsmath,amssymb,amsthm}
\usepackage{multirow}
\usepackage{colortbl}
\usepackage{booktabs}
\usepackage{amsmath}
\usepackage[table]{xcolor}
\definecolor{highlightcolor}{gray}{0.85}
\usepackage{tabularx}
\usepackage{adjustbox}
\usepackage{nicematrix}
\usepackage{tabularx}
\usepackage{changepage}
\usepackage{comment}
\usepackage{mathtools}
\usepackage{subfig}
\makeatletter
\newcommand{\mytag}[2]{%
  \text{#1}%
  \@bsphack
  \protected@write\@auxout{}%
         {\string\newlabel{#2}{{#1}{\thepage}}}%
  \@esphack
}
\makeatother
\usepackage{comment}
\usepackage{cancel}

\newtheoremstyle{examplesty}         
            {}                   
            {}                   
            {\upshape}           
            {}                   
            {\bfseries} 
            {}                   
            {1em}                
            {}                   

\theoremstyle{examplesty}

\usepackage{url}

\usepackage{algorithm}
\usepackage{algorithmic}

\usepackage{color}

\begin{document}
%
\title{BayesJudge: Bayesian Kernel Language Modelling with Confidence Uncertainty in Legal Judgment Prediction}
%
%
%
%

\author{Ubaid Azam,
        Imran Razzak,
        Shelly Vishwakarma,
        Hakim Hacid,\\
        Dell Zhang, 
        and Shoaib Jameel
\IEEEcompsocitemizethanks{\IEEEcompsocthanksitem 
U. Azam, S. Jameel and S. Vishwakarma are with School of electronics and Computer Science, University of Southampton, United Kindom. I. Razzak is with School of Computer Science and Engineering at University of New South Wales, Sydney, Australia. D. Zhang is with Thomson Reuters Labs, UK.
H. Hacid is with the Technology Innovation Institute, UAE.\protect\\
E-mail: {\it u.azam@soton.ac.uk} (corresponding author)}
\thanks{Manuscript received XXX YY, ZZZZ; revised XXX YY, ZZZZ.}}

%
%

\markboth{Journal of \LaTeX\ Class Files,~Vol.~14, No.~8, August~2015}%
{Shell \MakeLowercase{\textit{et al.}}: Bare Advanced Demo of IEEEtran.cls for IEEE Computer Society Journals}
%



\IEEEtitleabstractindextext{%
\begin{abstract}
Predicting legal judgments with reliable confidence is paramount for responsible legal AI applications. While transformer-based deep neural networks (DNNs) like BERT have demonstrated promise in legal tasks, accurately assessing their prediction confidence remains crucial. We present a novel Bayesian approach called BayesJudge that harnesses the synergy between deep learning and deep Gaussian Processes to quantify uncertainty through Bayesian kernel Monte Carlo dropout. Our method leverages informative priors and flexible data modelling via kernels, surpassing existing methods in both predictive accuracy and confidence estimation as indicated through brier score. Extensive evaluations of public legal datasets showcase our model's superior performance across diverse tasks. We also introduce an optimal solution to automate the scrutiny of unreliable predictions, resulting in a significant increase in the accuracy of the model's predictions by up to 27\%. By empowering judges and legal professionals with more reliable information, our work paves the way for trustworthy and transparent legal AI applications that facilitate informed decisions grounded in both knowledge and quantified uncertainty.

\end{abstract}

\begin{IEEEkeywords}
Legal Judgement Prediction, Language models, Kernel methods, Reliability.
\end{IEEEkeywords}}

\maketitle
\IEEEdisplaynontitleabstractindextext

%
\IEEEpeerreviewmaketitle

\ifCLASSOPTIONcompsoc
\IEEEraisesectionheading{\section{Introduction}\label{sec:introduction}}
\else

\section{Introduction}
\label{sec:introduction}
\fi

\IEEEPARstart {P}{roblem:} \textcolor{blue}{“Alice, a legal professional at a law firm handles legal cases. To automate the legal judgment prediction process, her firm opted to utilize powerful transformer models, specifically Legal BERT and Legal RoBERTa. For that they collected a dataset of legal judgments encompassing various case types, including sentence overruling, violations, and plea judgments. Alice trained the model on the dataset and implemented it. However, upon receiving predictions, she became uncertain about potential errors, recognizing the significant consequences of inaccuracies in legal proceedings. Feeling a sense of urgency, she began manually evaluating the predictions, consuming more time than ever. She expressed a desire for a process that gauged the model's confidence level, allowing her to focus on scrutinizing cases where the model displayed lower certainty, rather than reviewing each case individually.”}

The legal domain comprises the guidelines, rules, and institutions responsible for implementing and upholding laws to maintain societal peace \cite{dong2021legal}. It involves creating, clarifying, and applying laws to maintain order, resolve disputes, and protect individual rights \cite{schauer2004limited}.  Expert legal professionals, well-versed in the technicalities of each case, manage legal proceedings. However, the legal system faces significant challenges due to substantial case backlogs \cite{raaijmakers2019artificial, sansone2022legal}.  For instance, India struggles with a staggering 44 million pending cases\footnote{H. C .N. J. Data, ``National judicial data grid (district and Taluka courts of India),'' 2023. }, while Louisiana attorneys handle an average of 50 cases daily, dedicating only 1-5 minutes per case to preparation \cite{oppel2019one}. Similarly, the UK's Crown Court had 62,766 cases awaiting trial as of September 2022\footnote{https://www.gov.uk/government/statistics/criminal-court-statistics-quarterly-july-to-september-2022/criminal-court-statistics-quarterly-july-to-september-2022}. This high demand for legal assistance, coupled with the limited number of legal experts, leads to several societal problems, particularly the lack of accessible legal support for low-income citizens \cite{legal2017justice}.

The fundamental challenge to tackling this issue lies in overcoming the legal system's heavy reliance on manual processes. An automated system is therefore essential to expedite these tasks and free up valuable human resources. This paper introduces a novel computational model that effectively and efficiently communicates prediction confidence to legal experts. This will empower them to prioritize their time on a smaller subset of cases with low confidence, ultimately improving the efficiency and accessibility of the legal system.

Quantifying uncertainty \cite{geng2023survey} in machine learning \cite{shelmanov2021certain, pei2022transformer} model predictions allows legal professionals and the public to understand the model's reasoning better. This fosters trust in the use of AI in legal processes and helps to identify potential biases or limitations within the model. Legal cases often involve nuanced and complex situations where clear-cut answers are not always available. In such cases, a model's confidence estimates can be invaluable \cite{vazhentsev2023hybrid, fadeeva2023lm}. By highlighting situations where uncertainty is high, these estimates can prompt further investigation or human intervention. This ensures that complex cases receive the necessary attention and prevents potentially unjust outcomes based on unreliable model predictions. Additionally, machine learning models can inadvertently inherit biases from the data they are trained on. Quantifying uncertainty can help identify areas where such biases may be present, informing efforts to mitigate their impact. This is crucial for ensuring the fair and just application of AI in the legal system. Ultimately, legal decisions are made by humans. Understanding a model's confidence level allows judges and lawyers to weigh the model's predictions alongside their expertise and judgment. This collaborative approach can lead to more effective and nuanced legal decision-making.

Several compelling examples illustrate the importance of quantifying uncertainty in legal AI applications. In bail decisions, a model with low confidence in its risk prediction may encourage a judge to set lower bail, striking a balance between individual liberty and public safety. Conversely, a model consistently expressing high uncertainty in cases involving certain demographics could flag potential bias, prompting an investigation into its training data and algorithms. Moreover, cases with high uncertainty scores could be prioritized for further investigation or expert review, ensuring complex cases receive the necessary attention. Quantifying predictive uncertainty in the legal domain through machine learning is, therefore, not simply a technical challenge; it is a crucial step towards ensuring the responsible, fair, and transparent application of AI in legal systems. By empowering legal professionals to make informed decisions, enhance transparency, and mitigate bias, it ultimately paves the way for a more just and effective legal system.

Machine learning has played a pivotal role in boosting efficiency and accelerating progress in various fields, as evidenced by \cite{angra2017machine, dargan2020survey}. The legal field has also undergone a significant transformation in recent years, fueled by the emergence of innovative transformer models like Legal-Bert \cite{chalkidis2020legal} and Custom Legal-Bert \cite{zheng2021does}. One key focus has been advancing the prediction of legal judgments \cite{cui2023survey}. However, while considerable effort has been invested in automating legal judgment prediction, there is a notable lack of publicly accessible datasets for fine-tuning models, particularly those focusing on few-shot learning \cite{hu2018few}. This data scarcity often necessitates training models in few-shot and few-label settings \cite{li2020slbcnn}. Under such circumstances, placing complete trust in the model's predictions becomes difficult, highlighting the critical importance of ensuring the reliability and dependability of model outputs.

While existing works have developed methods for quantifying prediction uncertainty in models, they fall short of harnessing the full potential of a Bayesian approach with priors and kernel methods. For instance, Miok et al., \cite{miok2022ban} utilize Monte Carlo dropout to estimate uncertainty in hate speech prediction, building upon the dropout mechanism introduced by \cite{gal2016dropout}. However, their model lacks a fully Bayesian framework \cite{folgoc2021mc}, which restricts the ability to incorporate prior knowledge. Incorporating priors offers several benefits. Firstly, priors enable us to inject existing knowledge into the analysis, particularly valuable when data is scarce or ambiguous. Secondly, they act as regularizers, preventing conclusions from being unduly influenced by noise. Another key advantage of BayesJudge lies in its utilization of popular kernel methods, which facilitate reliable data modelling. Most importantly, priors enable the quantification of uncertainty in estimates and predictions, crucial for understanding model limitations and making informed decisions.

This paper tackles a major concern in the legal domain \cite{surden2014machine}: accurately measuring and improving the confidence of machine learning models in their legal judgment predictions \cite{zheng2021does, biran2017explanation}. We address this by proposing a novel Bayesian Monte Carlo dropout \cite{gal2016dropout, islam2023certain} mechanism that reliably quantifies the classifier's estimation confidence. Our Bayesian approach demonstrably enhances confidence estimates \cite{erlygin2023uncertainty}, reducing the need for manual intervention and fostering trust in the model's predictions. Real-world legal data often exhibits intricate, non-linear relationships that conventional linear models struggle to capture. To overcome this, we leverage kernel methods. These methods implicitly map data into a higher-dimensional space where linear relationships emerge, allowing us to capture nuances and complexities invisible in the original space. Extensive experiments showcase that BayesJudge significantly outperforms existing approaches in its ability to accurately estimate confidence.

Our key contributions are the following:
\begin{itemize}
\item To address the issues surrounding a black-box model's shaky predictions, we introduce BayesJudge, a novel Bayesian approach to Monte Carlo Dropout in language modelling that tackles uncertainty head-on. BayesJudge incorporates prior knowledge and leverages powerful kernel methods for faithful data modelling, leading to more accurate and reliable predictions.
\item BayesJudge can faithfully identify risky predictions before they cause harm \cite{mosley2023perceptions}.
\item We have conducted a thorough quantitative and qualitative experimental analysis demonstrating the strengths of BayesJudge, especially in resource-constrained environments.
\item We have developed a methodology to model the ``optimal solution'' derived from our quantitative results, outlining a pathway to strengthen the confidence of uncertain predictions made by the model.
\end{itemize}

\section{Related work}
\label{sec:Related_work}
Legal judgment prediction (LJP) \cite{zhong2018legal, cui2023survey, medvedeva2023rethinking, ganguly2023legal} is a rapidly growing field within artificial intelligence and law \cite{zhang2023contrastive, modi2023semeval}, aiming to automatically predict the outcome of legal cases based on their textual descriptions. Anticipating legal decisions has captivated interest for several years \cite{lawlor1963computers}, and the emergence of transformer models has indeed transformed the realm of legal judgment prediction tasks \cite{valvoda2023role, chang2023survey, collenette2023explainable}. Numerous contemporary transformer models have been developed with specialized training on legal datasets to effectively tackle tasks within the legal domain \cite{chalkidis2020legal}. In recent times, there has been a considerable effort to enhance the effectiveness and precision of legal judgment prediction using diverse deep learning methodologies \cite{zhong2020iteratively, paul2020automatic, ge2021learning}.

A study by \cite{xu2020distinguish} stands out for its utilization of graph neural network to distinguish the confusing law articles for legal judgement prediction, a similar approach is developed by \cite{dong2021legal}. They devised an approach that conceptualizes the problem akin to a node classification challenge, where each node's label distribution depends on the characteristics of its graph neighbours. This strategy aims to achieve local consistency through relational learning. The experimental findings indicate notable performance improvements, particularly on benchmark datasets. Similarly, \cite{liu2023ml} employs a graph attention network technique to comprehend various law articles, facilitating their prediction within the framework of multi-law-aware legal judgment prediction.

Typically, predictions are derived from the textual input, often summarized through the judge's narrative. In \cite{ma2021legal}, the authors conducted legal judgment predictions in real court scenarios, encompassing not only the plaintiff's claim but also the court debate. Leveraging a multi-task learning approach, they derived the final predictions. Correspondingly, \cite{semo2022classactionprediction} highlights the same concern and conducted Legal Judgment Prediction (LJP) on a more practical dataset, comprising plaintiffs' complaints as input instead of the court-written fact summaries. Their study concluded that undertaking legal judgment prediction in a real-world scenario, with plaintiffs' data as a direct input, poses increased difficulty.

Kevin in \cite{ashley2019brief}, provides a historical overview of LJP research, tracing its roots back to the early days of artificial intelligence. It highlights the evolving goals and challenges of LJP over time. The work by \cite{xu-etal-2020-distinguish} tackles the challenge of distinguishing between similar yet legally distinct articles in LJP tasks. They propose a novel graph neural network architecture called LADAN to automatically learn subtle differences between confusing law articles, leading to more accurate predictions. In \cite{ma2021legal}, the authors investigate LJP in a real-world court setting, focusing on criminal cases. They propose a multi-stage learning framework that leverages different types of legal data, including legal documents, court debates, and judicial facts, to improve prediction accuracy.

In \cite{chien2024legal}, the authors compare different machine learning models for judgment prediction and highlight the benefits of using indictments over verdicts. Furthermore, it emphasizes the importance of error analysis in identifying and rectifying weaknesses in indictments, ultimately leading to stronger prosecutions. In \cite{wang2024graph}, the authors introduce GraSCL, a framework that incorporates both graph reasoning and supervised contrastive learning (SCL) techniques. GraSCL offers a promising approach to LJP by explicitly modelling the relationships between case elements and leveraging label dependencies within a unified framework. In \cite{le2024topology}, the authors introduce a promising approach for civil case judgment prediction by considering subtask dependencies and multiple perspectives. This work can potentially help legal professionals, judicial systems, and researchers understand and predict court outcomes more accurately. In \cite{jacob2022using}, the authors developed a study that applies deep learning models to predict appeal outcomes in Brazilian federal courts based on real case data. This data includes text from rulings and other relevant documents. The study suggests that deep learning can be a valuable tool for legal professionals. It can help lawyers and judges make informed decisions about whether to pursue an appeal, potentially streamlining the judicial process and reducing frivolous appeals.

In \cite{deng2023syllogistic}, the authors developed a model that studies the use of syllogistic reasoning (a specific form of deductive logic) for analyzing legal judgments and argues for its importance in developing trusted legal judgment assistants using LLMs. While the paper focuses on criminal law cases, the framework could be extended to other legal domains. LLMs show moderate accuracy in identifying major and minor premises but struggle with generating high-quality legal conclusions. In \cite{zhang2022interpretable}, the authors developed a new model to be more interpretable than traditional models, which are often black boxes that provide predictions without explanation. The prediction of the prison term is made using the conditional probability distribution (CPD) table of the pooling of the two representations.

Predominantly, research within the legal domain leverages deep neural networks due to their effectiveness \cite{cui2023survey}. However, a common limitation of these models lies in their inability to quantify their output with confidence. To address this, \cite{gal2016dropout} uses dropout as a Bayesian approximation, representing uncertainty in deep learning models. Another approach based on deep ensembles proposed by \cite{lakshminarayanan2017simple} to estimate prediction uncertainty. A simple and general-purpose approach named SWAG has been introduced by \cite{maddox2019simple} for effectively representing and calibrating uncertainty in the realm of deep learning.

Deep learning and Gaussian Processes (GPs) \cite{williams2006gaussian} are both powerful machine learning approaches \cite{lee2017deep, damianou2013deep}, sharing similarities but also exhibiting key differences. Both utilize a layered architecture, each layer extracting increasingly complex features from the data, enabling them to handle intricate relationships and patterns. Furthermore, both can model non-linear relationships between inputs and outputs, crucial for real-world problems where data complexity reigns. Finally, both learn from data by adjusting internal parameters and adapting to diverse tasks and datasets. For interpretability and uncertainty quantification, however, GPs may be the preferred choice.

To the best of our knowledge, there has been no prior exploration in the field of legal judgment prediction aimed at addressing the uncertainty inherent in these predictions, a matter of substantial importance as inaccurate predictions can have severe consequences. This concern is underscored by \cite{carlsson2023legal}, who questions the reliability of automated decision-making processes. With the introduction of our proposed Bayesjudge model, we are equipped to discern predictions that carry uncertainty, prompting closer examination and scrutiny by legal experts.

\section{Our Novel BayesJudge Model}
\label{sec:Method}
This section investigates the technical details of the BayesJudge model, which provides more reliable uncertainty quantification for its predictive outputs. This uncertainty stems from the inherent lack of complete knowledge about the true relationship between input and output data. The BayesJudge model addresses this uncertainty by distinguishing itself from the standard Monte Carlo dropout model \cite{gal2016dropout, magris2023bayesian} in two key ways.

BayesJudge leverages a kernel function's inherent advantages, offering a diverse range of choices tailored to specific data types and problems. This flexibility empowers one to select the kernel that best suits the task and data, granting greater control over the learning process. Additionally, kernels like the Gaussian possess built-in regularization properties. This mitigates overfitting, a common issue in machine learning where models memorize training data instead of generalizing to unseen examples. While standard dropout tackles overfitting, BayesJudge further strengthens this safeguard through the chosen kernel function.

Our second innovation leverages the power of priors within the Monte Carlo dropout framework proposed by \cite{shelmanov2021certain}. Priors offer vital flexibility by enabling the integration of our prior knowledge and beliefs about the problem domain. This strategic injection of information steers the model towards solutions that resonate with real-world expectations. Furthermore, by incorporating priors, we imbue our Bayesian models with the ability to quantify uncertainty in their predictions. This explicit uncertainty estimation is indispensable for tasks where comprehending the model's limitations is paramount, such as medical diagnosis or legal prediction. Finally, priors facilitate the robust comparison and selection of different models based on their posterior probabilities. This data-driven approach empowers us to identify the model that aligns most effectively with both the observed data and our prior beliefs.

We first define the notations that we will use in this paper. We define \(\mathbf{\hat{\vartheta}}\) as the output of the model. The model comprises \(L\) layers. We denote the loss function such as softmax loss as \(\mathcal{L}(.,.)\). The weight matrices are denoted by \(\Theta\) with the dimension \(\digamma_i \times \digamma_{i-1}\). As always, there is a bias in the model that we denote as \(\mathbf{b}_i\) whose dimension is \(\digamma_i\) for each layer \(i=1,2,\cdots,L\). The observed variable is denoted by \(\vartheta\) for the corresponding input \(\mathbf{\pi}_i\) for \(1 \le i \le D\) data points. The input and the outputs are denoted by \(\mathbf{\Pi},\mathbf{\Omega}\). The regularisation parameter in \(L_2\) regularisation is denoted by \(\psi\). The weight matrices are denoted by \(\Phi\). The binary vectors are denoted by \(\zeta\). The vector dimension is denoted by \(R\).

Imagine a neural network constantly questioning its strength. It throws curveballs at itself during training, randomly deactivating neurons and forcing others to rise to the occasion. This rigorous process, known as Monte Carlo dropout, builds a network that's less likely to overfit, easily adapts to new data, and even acknowledges its uncertainty. For every input, the network performs a random draw for each neuron in its hidden layers. With a layer-specific probability \(p_i\), a neuron is temporarily silenced, demanding that others compensate. The network adjusts its weights and biases based on these modified inputs, learning to rely on diverse combinations of neurons instead of becoming overly reliant on specific ones. During backpropagation, the network leverages the same silencing pattern, ensuring consistent learning and preventing conflicting signals from the deactivated neurons. As demonstrated in \cite{gal2016dropout} there is a connection between the commonly used technique of dropout in deep neural networks and Bayesian inference in Gaussian processes \cite{khan2019approximate, mackay1998introduction}. The authors essentially explain how dropout, often understood as a regularizer to prevent overfitting, can be interpreted as a way to represent and quantify the uncertainty of a deep learning model.

Instead of relying on a single, deterministic prediction, Monte Carlo dropout employs randomness to provide a probabilistic view of the model's output. During training, dropout is applied as usual, but it remains active even at test time. This means that the network configuration changes with each pass, as random nodes/links are kept or dropped. Consequently, the prediction for a given data point becomes non-deterministic. This variability reflects the model's uncertainty around the prediction and allows us to interpret the outputs as samples from a probabilistic distribution. Consider, for example, running a sentiment analysis model with Monte Carlo dropout on the phrase ``the movie was underwhelming.'' The model might assign a negative sentiment 80\% of the time and a neutral 20\%, capturing the nuanced uncertainty in the statement.

An approximate predictive distribution, in the context of statistics and probability, refers to a probability distribution that is used to estimate the likelihood of future observations based on current data and a model, but with the acknowledgement that it might not be entirely accurate. When the kernel function is applied to the input vectors, let \(\kappa(\mathbf{\pi})\) denote the mapped feature vectors. Given the weight matrices \(\mathbf{M}_i\) of dimension \(K_i \times K_{i-1}\), bias vectors \(\mathbf{b}_i\) of dimensions \(K_i\), and binary vectors \(\mathbf{z}_i\) of dimensions \(K_{i-1}\) for each layer \(i=1,\cdots,L\), as well as the approximating variational distribution:

\begin{equation}
    q(\mathbf{\vartheta}^*|\mathbf{\pi}^*) = \int p(\mathbf{\vartheta}^*|\mathbf{\pi}^*, \mathbf{\{M}_i\}_{i=1}^L)q(\mathbf{\{M}_i\}_{i=1}^L)d\mathbf{\{M}_i\}_{i=1}^L
\end{equation}

\noindent The equation above can be expressed in the following form:


\begin{equation}
\begin{split}
    q(\mathbf{\vartheta}^*|\mathbf{\pi}^*) := & \mathcal{N}\big(\mathbf{\vartheta}^*;\hat{\mathbf{\vartheta}}^*(\kappa(\mathbf{\pi}),\mathbf{\zeta}_1,\cdots,\mathbf{\zeta}_L), \\
    & \tau^{-1}\mathbf{I}_R\big) \text{Bern}(\mathbf{\zeta}_1),\cdots,\text{Bern}(\mathbf{\zeta}_L)
\end{split}
\end{equation}

\noindent for some \(\tau>0\), with


\begin{equation}
\begin{split}
    \hat{\mathbf{y}}^{*} = & \sqrt{\frac{1}{\digamma_L}}(\mathbf{\Phi}_L\mathbf{\zeta}_L)\sigma\Big(\cdots\sqrt{\frac{1}{K_1}} \sqrt{\frac{1}{\digamma_1}} \\
    & 
    (\mathbf{\Phi}_2\mathbf{\zeta}_2 \sigma\big( (\mathbf{\Phi}_1 \mathbf{\zeta}_1 \big)\kappa(\mathbf{\pi})^*+\mathbf{b}_i) \cdots \Big)
\end{split}
\end{equation}

\noindent we have,

\begin{equation}
    \mathop{\mathbb{E}_{q(\mathbf{\vartheta}^*|\kappa(\mathbf{\pi})^*)}}(\mathbf{\vartheta}^*) \approx \frac{1}{T}\sum_{1}^{T}\hat{\mathbf{\vartheta}}^*(\kappa(\mathbf{\pi})^*,\hat{\mathbf{\zeta}}_{1,t},\cdots,\hat{\mathbf{\zeta}}_{L,t})
\end{equation}

\noindent with \(\hat{\mathbf{\zeta}}_{i,t} \sim \text{Bern}(p_i)\).

Since we have a prior distribution over \(p_i\), we write the expression as:

\begin{equation}
    p_i \sim \text{Beta}(\alpha,\beta), p_i \in [0,1]
\end{equation}

\noindent where \(\alpha\) and \(\beta\) are the parameters of the Beta distribution. The parameter \(\alpha\) represents the number of ``successes'' in a hypothetical experiment and \(\beta\) represents the number of ``failures'' in the same experiment. Note that the choice of the Beta distribution is mainly due to conjugacy \cite{fink1997compendium}. A conjugate prior is a special type of prior distribution used in Bayesian inference, where it has a unique and convenient property: when combined with the likelihood function of the observed data, the resulting posterior distribution also belongs to the same family of distributions as the prior. This makes working with conjugate priors in Bayesian analysis, particularly advantageous.

The posterior distribution can thus be represented as:

\begin{equation}
    P(p_i|\Pi) = \text{Beta}(\alpha_D,\beta_D)
\end{equation}

\noindent we can denote \(\alpha_D = \sum_{d=1}^{D} \pi_d+\alpha\) and \(\beta_D=D-\sum_{d=1}^{D}\pi_d+\beta\). Among the various kernel functions available, such as the radial basis kernel \cite{patle2013svm}, our experiments yielded the best results with the squared kernel. This choice was guided by the squared kernel's ease of differentiation and minimal computational overhead on the model.

\section{Experiments and Results}
In this section, we detail the evaluation methodology and models used in our experiments. We then compare our approach with different configurations of the comparative model proposed by \cite{miok2022ban}, which relies on the BERT-base model as its core component.

\subsection{Datasets}
We leverage two openly accessible popular datasets within the legal domain, namely the ECHR dataset \cite{zheng2021does} and the Overruling dataset \cite{chalkidis2019neural} for our experiments. The ECHR dataset comprises 11.5k cases, providing a detailed account of facts and any potential violations of articles. Each case is assigned an importance score. The dataset is divided into 7100 and 1380 cases for training and validation, respectively, with an additional 2998 cases allocated for testing. The labels are binary, aiming to determine whether a case involves a violation of any article. The Overruling dataset constitutes a binary classification task, determining whether a statement nullifies the precedent case order through a constitutionally valid statute or a decision by the same or higher-ranking court, which establishes a distinct rule regarding the point of law in question. This dataset encompasses a total of 2400 statements.

\subsection{Comparative Models}
To evaluate the performance of pre-trained transformers when used in conjunction with BayesJudge in the legal domain, we used four models, three of which were specifically trained on legal data. This focus further examined the suitability of different models for legal tasks under various experimental setups. We compared the performance of models incorporating our technique with the baseline method of \cite{miok2022ban}, which originally used only MCD BERT. For a fair comparison, we implemented their technique with other legal domain transformer models as well. All models were fine-tuned for 100 epochs with the Adam optimizer (epsilon=1e-8, learning rate=2e-5) and early stopping based on validation loss. We describe the transformer models used to model the text below.

\noindent \textit{Bert-base-uncased}: Bert-base-uncased, introduced by \cite{devlin2018bert}, is pre-trained on a vast corpus of English data. This model comprises 110 million parameters, 12 heads, 768 dimensions, and 12 layers, each featuring 12 self-attention heads.

\noindent \textit{Legal-Bert-base-uncased}: Legal-Bert, developed by \cite{chalkidis2020legal}, was devised to facilitate NLP research within the legal domain. This model undergoes pre-training on a legal corpus encompassing 12 gigabytes of diverse legal text in English, spanning various fields such as legislation and contract cases. Notably, it shares the same number of parameters, dimensions, and heads as the Bert-base model.

\noindent \textit{Legal-Roberta-base}: Recently, in \cite{chalkidis2023lexfiles}, the authors introduced the Legal-Roberta model tailored for NLP applications in the legal domain. This model is specifically designed to enhance the training of legal-oriented language models. The presented model undergoes training with a novel tokenizer, incorporating 50,000 Byte Pair Encodings (BPEs), and is trained on a diverse LeXFiles corpus.

\noindent \textit{Custom-legalbert}: Zheng et al., \cite{zheng2021does} trained on an extensive corpus of 37 gigabytes, comprising legal decisions from both federal and state courts. The model incorporates a custom domain-specific legal vocabulary.

\subsection{Evaluation Methodology}
We conducted a thorough evaluation through two extensive sets of experiments. Initially, we scrutinized the behaviour of transformer models in a few-shot setting, acknowledging the limited availability of publicly accessible data in the legal domain \cite{hu2018few}. We conducted experiments under zero-shot, five-shot, fifteen-shot, and full data split scenarios. To measure the generalization performance of BayesJudge, we conducted experiments exploiting a 5-fold cross-validation strategy. To ensure a reliable comparison, we conduct multiple rounds of experiments, randomly selecting samples for the few-shot experiments. This iterative process is replicated five times, ensuring a thorough evaluation of the model's performance. This approach has been used in different Bayesian settings to ensure the reliability of the results \cite{zhu2009medlda, jameel2015unified}. We reported the mean values in our experimental results.

In the second set of experiments, we employed adversarial testing by paraphrasing the test dataset using the ChatGPT paraphraser proposed by \cite{vorobev2023paraphrasing}. This approach aimed to assess how the models respond when the test dataset structure is altered. Publicly available legal domain datasets often include technical terms \cite{medvedeva2023legal}, which may not align with the reality where complainants or litigants are often laypeople, introducing non-technical language. Performance across all tasks was assessed using precision, recall, accuracy, and F1 score.

Given that our proposed methodology enables transformer models to generate predictions in a probabilistic manner, we utilized the Brier Score to evaluate the calibration of the model's predicted probabilities. The Brier Score \cite{parlett2019exploring}, representing the mean squared difference between predicted probabilities and actual outcomes, ranges from 0 to 1, with 0 indicating a perfect match.

The Brier Score formula is provided in Equation~\ref{eq:brier_binary}  
\begin{equation}
    Brier\ Score = \frac{1}{D} \sum_{i=1}^{D} (P_i - O_i)^2 \label{eq:brier_binary}
\end{equation}

\noindent where, $D$ denotes the total number of instances, $P_i$ represents the predicted probability of the positive class for the \(i^{\text{th}}\) instance, and $O_i$ corresponds to the actual outcome (either 0 or 1) for the \(i^{\text{th}}\) instance.

\subsection{Experimental Results}
We present the results of our computational model alongside those of comparative models. We convincingly demonstrate that BayesJudge outperforms its counterparts in legal judgment prediction tasks. This comprehensive evaluation showcases BayesJudge's effectiveness from multiple perspectives. Firstly, we establish BayesJudge's consistent superiority over comparative models in predicting legal judgments. Secondly, we delve into how BayesJudge adeptly handles data uncertainties and reliably models them. We further substantiate this by conducting experiments involving input text modifications, such as paraphrasing, to illustrate how different models perform under varying settings and input types. Finally, we present qualitative results to complement the quantitative analysis, providing a holistic picture of BayesJudge's capabilities.

\subsubsection{Legal Judgment Prediction}
In Table~\ref{tab:overuliingtable}, we present the outcomes of various models on the Overruling dataset, showcasing their performance in zero, five, fifteen-shot, and full-data scenarios. The experiments were specifically designed to scrutinize how models respond to low-resource conditions. LJP presents a notable challenge concerning the availability and richness of datasets \cite{cui2023survey}, our experiments aim to shed light on the behaviour of transformer models in low-resource settings. Moreover, the introduction of our proposed Bayesjudge model is expected to assist in flagging less certain predictions, contributing to a more nuanced analysis.

In the full-data scenario, Legal-Bert emerges as the top performer, achieving an F1 score of 0.979 with a Brier score of 0.022, closely trailed by Custom Legal-Bert with an F1 score of 0.973 and a Brier score of 0.024. Contrastingly, in situations with limited resources for model training, Custom Legal-Bert outshines all other models, demonstrating a substantial difference in F1 scores. Notably, it attains F1 and Brier scores of 0.950 and 0.086 and 0.785 and 0.165 under the 15-shot and 5-shot scenarios, respectively, showcasing its effectiveness in low-resource settings. The Legal-bert performed second best and attained an F1 and Brier score of 0.886, 0.180, and 0.708, 0.225, respectively.

Legal-Roberta consistently demonstrated the weakest performance across all experiments, exhibiting poorer results than the baseline Bert model, which is not even specifically tailored for legal data. It recorded F1 and Brier scores of 0.919 and 0.047, 0.67 and 0.270, and 0.67 and 0.309 for full data split, 15-shot, and 5-shot scenarios, respectively. Particularly noteworthy is its tendency, under both 15-shot and zero-shot conditions, to predict only one class while entirely neglecting the other. In contrast, Bert showcased comparatively better performance, securing F1 and Brier scores of 0.965 and 0.032, 0.838 and 0.13, and 0.624 and 0.201 for full data split, 15-shot, and 5-shot scenarios, respectively.

Similar behaviour can be noticed in the ECHR dataset. Table ~\ref{tab:ECHRTable} describes the evaluation of BayesJudge on the ECHR dataset. Legal-Bert emerged as the top performer for both full-data and few-shot scenarios, boasting F1 and Brier scores of 0.820 and 0.139, 0.659 and 0.2, and 0.51 and 0.218 in the full-data split, 15-shot, and 5-shot scenarios, respectively. The second-strongest model was Custom-Legal Bert, particularly shining in few-shot scenarios with F1 and Brier scores of 0.485 and 0.245, and 0.452 and 0.242 for 15-shot and 5-shot scenarios, respectively. Much like the prior dataset, Legal-Roberta showcased suboptimal performance in the ECHR dataset under few-shot conditions, registering F1 scores of 0.397 for both 15-shot and 5-shot scenarios.

It is worth noting that that our proposed method exhibits a significant enhancement in both F1 and Brier scores when compared to the baseline model across all experiments on both datasets, particularly in resource-constrained scenarios such as the 5 and 15-shot settings. This underscores its effectiveness in improving classification performance and reliability.

\begin{table}[]
\caption{Overrulling Dataset Results}
\scalebox{1.05}
{
\begin{tabular}{ccccc>{\columncolor{highlightcolor}}c}
\toprule
                                       & \multicolumn{5}{l}{\textbf{Overrulling Dataset (80-20 split)}}                    \\
                                       
\multirow{-2}{*}{\textbf{Model Names}} & Pre            & Rec            & F1             & Acc            & Brier          \\
\toprule
\multicolumn{6}{c@{\hspace{-6em}}}{Baseline Models}                                                                   \\
\hline
BERT                                   & 0.971          & 0.959          & 0.965          & 0.965          & 0.040          \\
Custom Legal-BERT                      & 0.975 & 0.962          & 0.968          & 0.968          & 0.027          \\
\cellcolor[HTML]{FFFFFF}Legal-BERT     & 0.973          & 0.979 & 0.973 & 0.972 & 0.023 \\
Legal-RoBERTA                          & 0.987          & 0.789          & 0.878          & 0.889          & 0.104          \\

\hline

\multicolumn{6}{c@{\hspace{-6em}}}{BayesJudge Models}                                                                   \\
\hline
BERT                                   & 0.959          & 0.971          & 0.965          & 0.965          & 0.032          \\
Custom Legal-BERT                      & \textbf{0.987} & 0.959          & 0.973          & 0.973          & 0.024          \\
\cellcolor[HTML]{FFFFFF}Legal-BERT     & 0.983          & \textbf{0.975} & \textbf{0.979} & \textbf{0.979} & \textbf{0.022} \\
Legal-RoBERTA                          & 0.977          & 0.868          & 0.919          & 0.923          & 0.047          \\
\hline
\multicolumn{6}{c@{\hspace{-6em}}}{\textbf{Overrulling Dataset 15 Shot}}                                                                   \\

\hline
\multicolumn{6}{c@{\hspace{-6em}}}{Baseline Models}                                                                   \\
\hline
BERT                                   & 0.792          & 0.897          & 0.841          & 0.829          & 0.13           \\
Custom Legal-BERT                      & 0.893 & 0.966          & 0.928          & 0.925          & 0.092          \\
\cellcolor[HTML]{FFFFFF}Legal-BERT     & 0.581          & 0.991 & 0.732 & 0.635 & 0.200 \\
Legal-RoBERTA                          & 0.504          & 1          & 0.670          & 0.504          & 0.300          \\
\hline
\multicolumn{6}{c@{\hspace{-6em}}}{BayesJudge Models}                                                                   \\
\hline
BERT                                   & 0.893          & 0.789          & 0.838          & 0.846          & 0.13           \\
Custom Legal-BERT                      & \textbf{0.922} & 0.979          & \textbf{0.950} & \textbf{0.948} & \textbf{0.086} \\
\cellcolor[HTML]{FFFFFF}Legal-BERT     & 0.807          & 0.983          & 0.886          & 0.873          & 0.180          \\
Legal-RoBERTA                          & 0.504          & \textbf{1}     & 0.670          & 0.504          & 0.270          \\
\hline
\multicolumn{6}{c@{\hspace{-6em}}}{\textbf{Overrulling Dataset 5 Shot}}                                                                    \\

\hline
\multicolumn{6}{c@{\hspace{-6em}}}{Baseline Models}                                                                   \\
\hline
BERT                                    & 0.548          & 0.996          & 0.707          & 0.583          & 0.254          \\
Custom Legal-BERT                      & 0.642 & 0.963          & 0.770          & 0.710          & 0.178          \\
\cellcolor[HTML]{FFFFFF}Legal-BERT     & 0.509          & 1 & 0.675 & 0.514 & 0.254 \\
Legal-RoBERTA                          & 0.504          & 1          & 0.670          & 0.496          & 0.335          \\
\hline
\multicolumn{6}{c@{\hspace{-6em}}}{BayesJudge Models}                                                                   \\
\hline
BERT                                   & 0.892          & 0.479          & 0.624          & 0.708          & 0.201          \\
Custom Legal-BERT                      & \textbf{0.667} & 0.954          & \textbf{0.785} & \textbf{0.737} & \textbf{0.165} \\
\cellcolor[HTML]{FFFFFF}Legal-BERT     & 0.548          & 1              & 0.708          & 0.583          & 0.225          \\
Legal-RoBERTA                          & 0.504          & \textbf{1}     & 0.670          & 0.496          & 0.309          \\
\hline
\multicolumn{6}{c@{\hspace{-8em}}}{\textbf{Overrulling Dataset Zero Shot}}                                                                 \\
\hline
\multicolumn{6}{c@{\hspace{-6em}}}{Baseline Models}                                                                   \\
\hline
BERT                                   &0.504 & 0.988          & 0.668          & 0.504 & 0.255 \\
Custom Legal-BERT                      & 0.463 & 0.789          & 0.584          & 0.433          & 0.280          \\
\cellcolor[HTML]{FFFFFF}Legal-BERT     & 0.501          & 0.975 & 0.661 & 0.497 & 0.264 \\
Legal-RoBERTA                          & 0.504          & 0.972          & 0.670          & 0.496          & 0.255          \\
\hline
\multicolumn{6}{c@{\hspace{-6em}}}{BayesJudge Models}                                                                   \\
\hline
BERT                                   & \textbf{0.507} & 0.950          & 0.661          & \textbf{0.508} & \textbf{0.249} \\
Custom Legal-BERT                      & 0.503          & 0.967          & 0.653          & 0.496          & 0.250          \\
\cellcolor[HTML]{FFFFFF}Legal-BERT     & 0.504          & \textbf{0.99}  & \textbf{0.671} & 0.504          & 0.261          \\
Legal-RoBERTA                          & 0.504          & 0.972          & 0.670          & 0.496          & 0.252         \\ \hline
\end{tabular}
}
\label{tab:overuliingtable}
\end{table}

\begin{table}[]
\caption{ECHR Dataset Results}
\scalebox{1.05}
{
\begin{tabular}{ccccc>{\columncolor{highlightcolor}}c}
\toprule
                                       & \multicolumn{4}{c}{\textbf{ECHR Dataset (80-20 split)}}                                                                                                                   \\
\multirow{-2}{*}{\textbf{Model Names}} & Pre                                    & Rec                                    & F1             & Acc                                    & Brier                         \\
\toprule
\multicolumn{6}{c@{\hspace{-6em}}}{Baseline Models}                                                                   \\
\hline
BERT                                   & 0.829                                  & 0.808                                  & 0.817          & 0.840                                  & 0.138                         \\
Custom Legal-BERT                      & 0.823 & 0.803          & 0.811          & 0.835          & 0.149          \\
\cellcolor[HTML]{FFFFFF}Legal-BERT     & 0.836          & 0.808 & 0.819 & 0.843 & 0.140 \\
Legal-RoBERTA                          & 0.851          & 0.799          & 0.816          & 0.846          & 0.128          \\
\hline

\multicolumn{6}{c@{\hspace{-6em}}}{BayesJudge Models}                                                                   \\
\hline
BERT                                   & 0.831                                  & 0.810                                  & 0.819          & 0.842                                  & 0.137                         \\
Custom Legal-BERT                      & 0.826                                  & 0.795                                  & 0.807          & 0.833                                  & 0.145                         \\
\cellcolor[HTML]{FFFFFF}Legal-BERT     & 0.826                                  & \textbf{0.815}                         & \textbf{0.820} & 0.840                                  & 0.139                         \\
Legal-RoBERTA                          & \textbf{0.847}                         & 0.804                                  & 0.819          & \textbf{0.846}                         & \textbf{0.127}                \\
\hline
\multicolumn{6}{c@{\hspace{-4em}}}{\textbf{ECHR Dataset 15 Shot}}                                                                                                                                                                  \\
\hline
\multicolumn{6}{c@{\hspace{-6em}}}{Baseline Models}                                                                   \\
\hline
BERT                                   & 0.483                                  & 0.481                                  & 0.475          & 0.498                                  & 0.250                         \\
Custom Legal-BERT                      & 0.639 & 0.532          & 0.482          & 0.670          & 0.301          \\
\cellcolor[HTML]{FFFFFF}Legal-BERT     & 0.644          & 0.630 & 0.634 & 0.686 & 0.215 \\
Legal-RoBERTA                          & 0.506          & 0.502          & 0.448          & 0.636          & 0.240          \\
\hline
\multicolumn{6}{c@{\hspace{-6em}}}{BayesJudge Models}                                                                   \\
\hline
BERT                                   & 0.530                                  & 0.510                                  & 0.466          & 0.640                                  & 0.236                         \\
Custom Legal-BERT                      & 0.486                                  & 0.486                                  & 0.485          & 0.525                                  & 0.245                         \\
\cellcolor[HTML]{FFFFFF}Legal-BERT     & \textbf{0.657}                         & \textbf{0.662}                         & \textbf{0.659} & \textbf{0.687}                         & \textbf{0.200}                \\
Legal-RoBERTA                          & 0.329                                  & 0.500                                  & 0.397          & 0.658                                  & 0.237                         \\
\hline
\multicolumn{6}{c@{\hspace{-4em}}}{\textbf{ECHR Dataset 5 Shot}}                                                                                                                                                                   \\
\hline
\multicolumn{6}{c@{\hspace{-6em}}}{Baseline Models}                                                                   \\
\hline
BERT                                   & 0.353                                  & 0.349                                  & 0.351          & 0.407                                  & 0.288                         \\
Custom Legal-BERT                      & 0.439 & 0.410          & 0.450          & 0.501          & 0.287          \\
\cellcolor[HTML]{FFFFFF}Legal-BERT     & 0.510          & 0.483 & 0.497 & 0.570 & 0.232 \\
Legal-RoBERTA                          & 0.516          & 0.517          & 0.479          & 0.480          & 0.252          \\
\hline
\multicolumn{6}{c@{\hspace{-6em}}}{BayesJudge Models}                                                                   \\
\hline
BERT                                   & 0.472                                  & 0.469                                  & 0.463          & 0.482                                  & 0.272                         \\
Custom Legal-BERT                      & 0.455                                  & 0.467                                  & 0.452          & 0.563                                  & 0.242                         \\
\cellcolor[HTML]{FFFFFF}Legal-BERT     & \textbf{0.519}                         & \textbf{0.516}                         & \textbf{0.511} & 0.596                                  & \textbf{0.218}                \\
Legal-RoBERTA                          & 0.329                                  & 0.5                                    & 0.397          & \textbf{0.658}                         & 0.236                         \\
\hline
\multicolumn{6}{c@{\hspace{-5em}}}{\textbf{ECHR Dataset Zero Shot}}                                                                                                                                                                \\
\hline
\multicolumn{6}{c@{\hspace{-6em}}}{Baseline Models}                                                                   \\
\hline
BERT-Baseline                                   & 0.361                                  & 0.436                                  & 0.286          & 0.321                                  & 0.302 \\
Custom Legal-BERT                      & 0.462 & 0.468          & 0.460          & 0.550          & 0.248          \\
\cellcolor[HTML]{FFFFFF}Legal-BERT     & 0.458          & 0.464 & 0.388 & 0.392 & 0.276 \\
Legal-RoBERTA                          & 0.497          & 0.499          & 0.300          & 0.358          & 0.268          \\
\hline
\multicolumn{6}{c@{\hspace{-6em}}}{BayesJudge Models}                                                                   \\
\hline
BERT                                   & \textbf{0.555}                                  & \textbf{0.559}                                 & \textbf{0.521}          & 0.524                                  & 0.248 \\
Custom Legal-BERT                      & 0.372                                  & 0.481                                  & 0.395          & 0.629                                  & \textbf{0.236}                \\
\cellcolor[HTML]{FFFFFF}Legal-BERT     & 0.415                                  & 0.406                                  & 0.403          & 0.424                                  & 0.262                         \\
Legal-RoBERTA                          & \cellcolor[HTML]{FFFFFF}0.522 & \cellcolor[HTML]{FFFFFF}0.509 & 0.468 & \cellcolor[HTML]{FFFFFF}\textbf{0.635} & 0.247      \\ \hline                  
\end{tabular}
}
\label{tab:ECHRTable}
\end{table}

\subsubsection{Adversarial Experiments}
Publicly accessible legal datasets often contain legal terminology and language, but real-world legal communication may differ significantly. Consider a passage from the dataset: ``to the extent that the Huffman opinion may be in conflict herewith, it is overruled.'' This includes key legal terms like ``conflict'' and ``overruled''. While training on the dataset helps the model learn these terms, it can also lead to overfitting. In real terms, the passage might be expressed as ``In case where Huffman opinion disagrees with this, hereby we don't accept it.''

To assess how legal models respond to datasets that better reflect real-world language, we trained the model on the original data and paraphrased only the test dataset. For example, a passage like ``for the reasons that follow, we approve the first district in the instant case and disapprove the decisions of the fourth district'' was paraphrased to ``For the causes explained, we agree with the first district in this case and disagree with the decisions of the fourth district.''

\begin{figure}[t]
    \centering
    \includegraphics[scale=0.5]{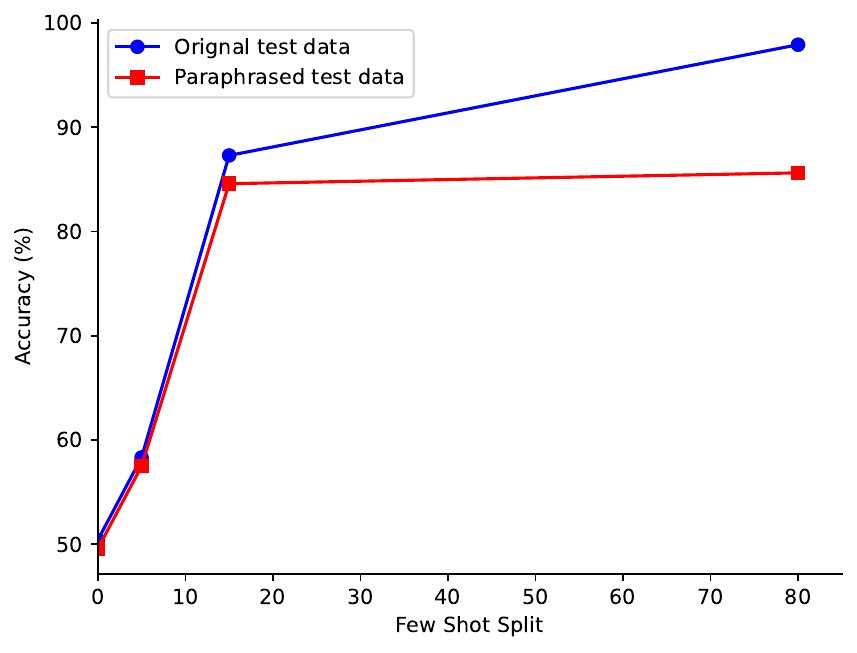}
    \caption{Accuracy comparison for Original vs Paraphrased test data results. We conducted simple paraphrasing to demonstrate how models will react to the real world when the text is simple. We showed results decreased as the text became simple which is expected.}
    \label{fig:accuracy_comparison}
\end{figure}

Paraphrasing the test data led to confusion in the model's predictions, as evidenced by drops in F1 scores across various transformer models, detailed in Table ~\ref{tab:overuliingtable_Paraphrased}. Figure~\ref{fig:accuracy_comparison} illustrates the accuracy comparison between the original and paraphrased test data on the Legal-Bert model.
Due to resource constraints, e.g., computational and limits imposed by OpenAI's freely available Application Programming Interface (API), we limited our adversarial experiments to the Overruling dataset exclusively.

It's crucial to emphasize that despite a decrease in F1 score after paraphrasing the test data, our BayesJudge Models still outperform the baseline approach, exhibiting higher F1 scores and lower Brier scores. This suggests improved classification performance and enhanced reliability compared to baseline models. Therefore, our proposed approach holds promise for making substantial contributions to improving prediction performance and reliability in real-world legal scenarios.

\begin{table}[]
\caption{Overrulling Dataset Paraphrased (denoted as Paraph) Results.}
\scalebox{1.05}
{
\begin{tabular}{ccccc>{\columncolor{highlightcolor}}c}
\toprule
                                       & \multicolumn{5}{c}{\textbf{Overrulling Dataset (80-20 split) Paraph}}       \\
\multirow{-2}{*}{\textbf{Model Names}} & Pre            & Rec         & F1             & Acc            & Brier          \\
\toprule
\multicolumn{6}{c@{\hspace{-6em}}}{Baseline Models}                                                                   \\
\hline
BERT                                   & 0.929          & 0.707          & 0.803          & 0.825          & 0.152          \\
Custom Legal-BERT                      & 0.941 & 0.785          & 0.856          & 0.867          & 0.116          \\
\cellcolor[HTML]{FFFFFF}Legal-BERT     & 0.983          & 0.727 & 0.836 & 0.856 & 0.125 \\
Legal-RoBERTA                          & 0.959          & 0.388          & 0.552          & 0.683          & 0.288          \\
\hline

\multicolumn{6}{c@{\hspace{-6em}}}{BayesJudge Models}                                                                   \\
\hline
BERT                                   & 0.934          & 0.822          & 0.875          & 0.881          & 0.112         \\
Custom Legal-BERT                      & 0.959          & 0.780 & 0.861 & 0.872 & 0.115 \\
\cellcolor[HTML]{FFFFFF}Legal-BERT     & \textbf{0.909} & \textbf{0.913}          & \textbf{0.911}          & \textbf{0.911}          & \textbf{0.073}          \\
Legal-RoBERTA                          & 0.869          & 0.740          & 0.799          & 0.813          & 0.164          \\
\hline
\multicolumn{6}{c@{\hspace{-7em}}}{\textbf{Overrulling Dataset 15 Shot Paraph}}                                                      \\
\hline
\multicolumn{6}{c@{\hspace{-6em}}}{Baseline Models}                                                                   \\
\hline
BERT                                   & 0.651         & 0.934          & 0.767          & 0.715          & 0.195          \\
Custom Legal-BERT                      & 0.745 & 0.921          & 0.824          & 0.802          & 0.151          \\
\cellcolor[HTML]{FFFFFF}Legal-BERT     & 0.726          & 0.747 & 0.737 & 0.731 & 0.199 \\
Legal-RoBERTA                          & 0.504          & 1          & 0.670          & 0.504          & 0.282          \\
\hline
\multicolumn{6}{c@{\hspace{-6em}}}{BayesJudge Models}                                                                   \\
\hline
BERT                                   & 0.747         & 0.781          & 0.764          & 0.756          & 0.172          \\
Custom Legal-BERT                      & 0.859          & 0.806          & 0.832          & 0.835          & \textbf{0.138}          \\
\cellcolor[HTML]{FFFFFF}Legal-BERT     & \textbf{0.908} & 0.773          & \textbf{0.835} & \textbf{0.846} & 0.166 \\
Legal-RoBERTA                          & 0.504          & \textbf{1} & 0.670          & 0.504          & 0.280          \\
\hline
\multicolumn{6}{c@{\hspace{-7em}}}{\textbf{Overrulling Dataset 5 Shot Paraph}}                                                       \\
\hline
\multicolumn{6}{c@{\hspace{-6em}}}{Baseline Models}                                                                   \\
\hline
BERT                                   & 0.519          & 1          & 0.684          & 0.533          & 0.242          \\
Custom Legal-BERT                      & 0.582 & 0.950          & 0.724          & 0.635          & 0.208          \\
\cellcolor[HTML]{FFFFFF}Legal-BERT     & 0.504          & 1 & 0.670 & 0.504 & 0.271 \\
Legal-RoBERTA                          & 0.504          & 0.980          & 0.665          & 0.495          & 0.314          \\
\hline
\multicolumn{6}{c@{\hspace{-6em}}}{BayesJudge Models}                                                                   \\
\hline
BERT                                   & 0.663          & 0.698          & 0.68          & 0.669          & 0.218          \\
Custom Legal-BERT                      & \textbf{0.616} & 0.921          & \textbf{0.738} & \textbf{0.671} & \textbf{0.201} \\
\cellcolor[HTML]{FFFFFF}Legal-BERT     & 0.544          & 0.954          & 0.694          & 0.575          & 0.243          \\
Legal-RoBERTA                          & 0.504          & \textbf{1}     & 0.670          & 0.496          & 0.298          \\
\hline
\multicolumn{6}{c@{\hspace{-8em}}}{\textbf{Overrulling Dataset Zero Shot Paraph}}                                                     \\
\hline
\multicolumn{6}{c@{\hspace{-6em}}}{Baseline Models}                                                                   \\
\hline
BERT                                    & 0.505          & 1              & 0.671 & 0.506 & 0.252 \\
Custom Legal-BERT                      & \textbf{0.619} & 0.161          & 0.255          & \textbf{0.527}          & 0.258          \\
\cellcolor[HTML]{FFFFFF}Legal-BERT     & 0.505          & 1 & 0.671 & 0.495 & 0.276 \\
Legal-RoBERTA                          & 0.504          & 1          & 0.670          & 0.504          & 0.254          \\
\hline
\multicolumn{6}{c@{\hspace{-6em}}}{BayesJudge Models}                                                                   \\
\hline
BERT                                   & 0.505          & 1              & \textbf{0.671} & 0.506 & 0.251 \\
Custom Legal-BERT                      & 0.505 & 0.963          & 0.663          & 0.506          & \textbf{0.249}          \\
\cellcolor[HTML]{FFFFFF}Legal-BERT     & 0.504          & 1              & 0.670          & 0.496          & 0.265          \\
Legal-RoBERTA                          & 0.504          & \textbf{1}     & 0.670          & 0.504          & 0.252     \\ \hline     
\end{tabular}
}
\label{tab:overuliingtable_Paraphrased}
\end{table}

\subsubsection{BayesJudge: Modelling Uncertainty}

\begin{figure}
    \centering
   
    \includegraphics[width=0.5 \textwidth]{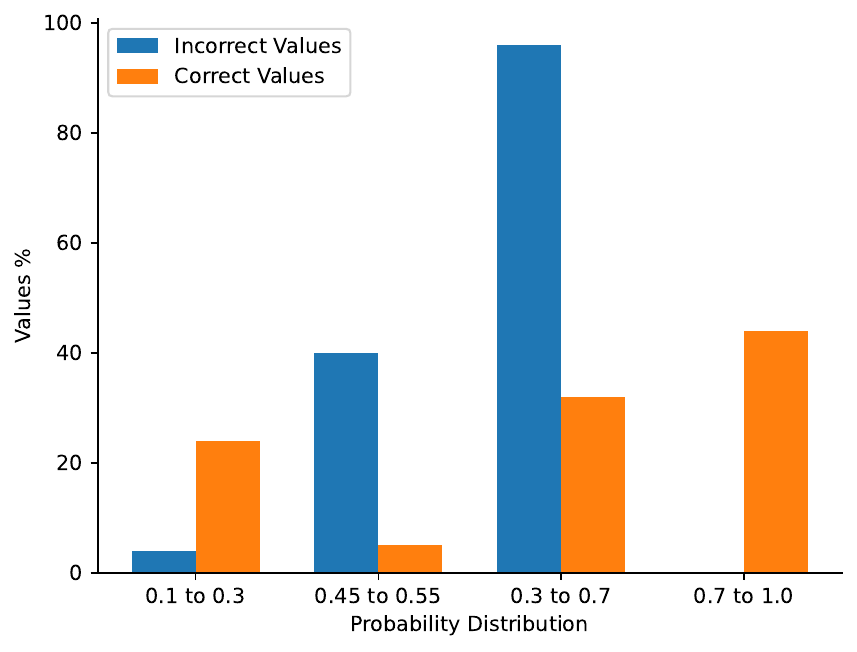}
    \caption{Probability distribution and prediction results for Custom Legal-BERT for 15 Shot.}
    \label{fig:Prediction Distribution}
\end{figure}

One of the key advantages of BayesJudge is its reliability in modelling uncertainty given its Bayesian architecture. We conduct a series of experiments demonstrating how BayesJudge can handle uncertainty in its predictions. To this end, we conduct error analysis and experiments surrounding modelling the confidence estimates learnt by the model.

Our BayesJudge model enables us to do error analysis by pinpointing predictions where the model encounters confusion and uncertainty in its decision-making. In Figure ~\ref{fig:Prediction Distribution}, we present the probability distribution of correct and incorrect predictions generated by the Custom Legal-BERT model. In the context of binary class classification, the model exhibits higher confidence in proximity to its true labels, namely, 0 and 1. We have delineated the values across four segments of the distribution spectrum. Between 0.1 to 0.3 and 0.7 to 1.0 ranges, the model demonstrates confidence in its predictions. Conversely, the model exhibits uncertainty at the borderline, specifically in the range of 0.45 to 0.55, where it is less certain about its predictions. To offer a broader perspective, we have also considered the probability distribution between 0.3 to 0.7. The figure distinctly illustrates that within the intervals of 0.1 to 0.3 and 0.7 to 1.0, where the model exhibits the highest confidence in its predictions, it predominantly predicts the correct values. Conversely, in distributions representing the model's uncertainty, it tends to predict incorrect values, indicating a state of confusion. Utilizing BayesJudge, we can effectively identify these flagged predictions, highlighting instances that require closer scrutiny.

As BayesJudge highlights the uncertain predictions, contributing to enhanced effectiveness and efficiency, requiring less effort as attention can be concentrated solely on those predictions where the model expresses uncertainty. The subsequent stage involves a meticulous examination of these less certain predictions. While one solution involves manual reevaluation that takes time and manpower, our paper introduces an alternative and effective method for automated scrutiny of these less certain predictions.

We conducted a series of experiments aimed at bolstering the confidence of the model's predictions at points where uncertainty was observed. In legal datasets, the presence of legal terms significantly influenced the predictions. Upon in-depth analysis, we observed that the model exhibited uncertainty in instances where the language was legally ambiguous or less certain we called them ``Simple Text''. To deal with uncertain predictions, we employed ChatGPT to rephrase uncertain instances, introducing more legal terms while maintaining the contextual integrity of the text, we call the text generated from this approach ``Advanced Legal Text''. What we did here is to pick uncertain predictions and add more legal terms using ChatGPT. To obtain the advanced legal text, our prompt was \textit{``Improve the vocabulary of the passage in Legal Terms''}.

Additionally, we carried out another set of experiments focusing on uncertain predictions, tasking ChatGPT to rephrase instances using legal terminology and expanding the text accordingly we named the generated text ``Extended Legal text''.  To obtain the extended legal text, we used the following prompt: \textit{``Write the passage fully advanced in legal terms and extend the sentences''}. Table \ref{tab:chatgpttext} demonstrates a practical example illustrating the effectiveness of both techniques. Initially, the passage was inaccurately predicted as class 1, but through our proposed methods, the passage was enriched with legal terms, enabling the model to correctly predict its class. The reason why BayesJudge works ideally here is that We are infusing additional legal terminology, leveraging the pre-training of our models on legal datasets, thereby enhancing their confidence and predictive capabilities.

\begin{figure}[t]
    \centering
    \includegraphics[width=0.5\textwidth]{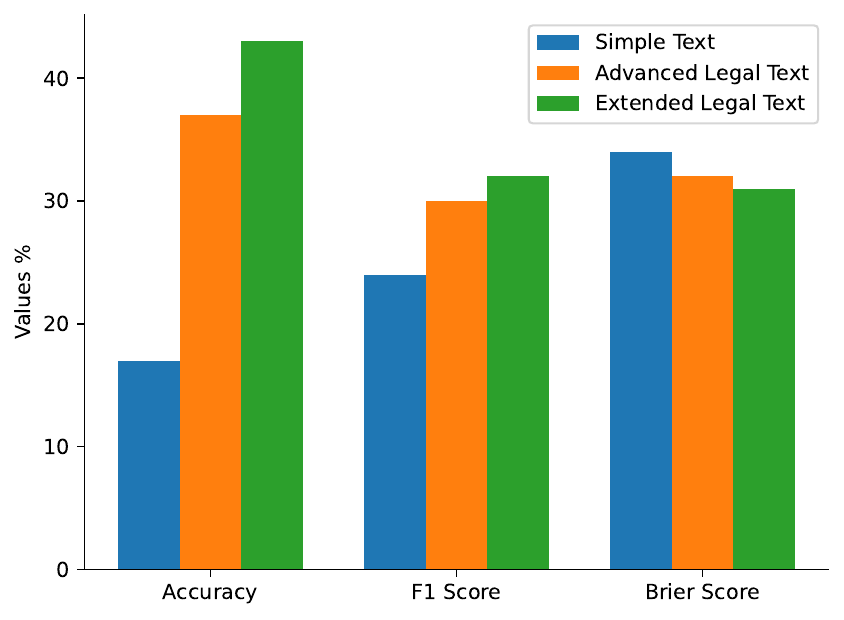}
    \caption{Accuracy and Brier score comparison of techniques on less certain predictions.}
    \label{fig:optimal solution}
\end{figure}

Figure~\ref{fig:optimal solution} provides a visual representation of the accuracy and Brier score of the aforementioned techniques applied to less certain predictions. These experiments were conducted using the Custom Legal-Bert model. Notably, our proposed optimal solution resulted in a 20\% increase in accuracy when leveraging advanced legal text and a 27\% improvement with the extended legal text technique. Furthermore, the Brier score decreased, indicating an enhancement in the model's confidence in its predictions compared to its previous state. As mentioned earlier, the reason why BayesJudge produces faithful results in this case is because we are injecting more legal terms and our models are pre-trained on legal datasets which helps to improve confidence and prediction.

\subsubsection{Kernel Functions and Beta Priors}
As previously mentioned, the squared kernel, a type of polynomial kernel with an exponent of 2, proved highly effective in our experiments. While we tested other kernels like Gaussian, linear, Laplacian, and sigmoid, none surpassed the results achieved with the squared kernel. All other kernels yielded results roughly 5\% lower than the squared kernel.

The squared kernel's success stems from its simplicity and ease of differentiation, unlike the linear kernel. Furthermore, it demonstrated computational efficiency compared to kernels like Gaussian and Laplacian.

We experimented with different values for the Beta priors, \(\alpha\) and \(\beta\). Ideally, we would automate the inference of these prior parameters by placing priors over them themselves, thereby achieving a fully Bayesian model through posterior inference on the hyperpriors. However, this approach incurs a significant computational burden. Fortunately, several Bayesian approaches, like topic models, have shown that fixing the prior values can lead to comparable results as those obtained through posterior inference on the priors \cite{blei2003latent, griffiths2003hierarchical, jameel2014latent}.

Experimenting with different values for the symmetric Beta priors, we found that setting them to 0.1 led to a 10\% decrease in the F1 score for legal judgment prediction. Further reducing the values to 0.001 yielded only marginally better results. The optimal configuration was achieved with symmetric Beta priors of \(\alpha\) = 0.0001 and \(\beta\) = 0.0001 which we use in all our experiments.

\begin{table}
\caption{Sample of Passage for each technique from Overruling dataset}
\scalebox{0.90}
{
\begin{tabular}{p{1.5cm}p{4cm}cc}
\hline
Type of text & Passage &Label & Pred Label \\
\hline
Simple & Alternatively, the state argues that there is no evidence that the police engaged in an "ask first, warn later" interrogation in violation of Seibert, and so at least his second statement, which he made after receiving Miranda warnings, was admissible. &0 & 1 \\
\hline
Advanced Legal & In the alternative, the state contends that there is an absence of evidence demonstrating that the police engaged in an 'ask first, warn later' interrogation, thereby violating Seibert. Hence, it argues that at least the defendant's second statement, made subsequent to receiving Miranda warnings, should be deemed admissible. & 0 & 0 \\
\hline
Extended Legal & In an alternative line of argument, the state posits that there exists an absence of evidence demonstrating that the police engaged in an 'ask first, warn later' interrogation, a practice purportedly in violation of Seibert. Consequently, the state contends that, at the very least, the defendant's second statement, made subsequent to receiving Miranda warnings, should be deemed admissible in accordance with established legal principles. & 0 & 0 \\
\hline
\end{tabular}
}
\label{tab:chatgpttext}
\end{table}

\section{Conclusions and Future Work}
The legal domain is undergoing a transformative shift, propelled by the arrival of sophisticated language models, particularly those with vast capacity. While challenges remain, this disruption also presents enticing opportunities. Language models can empower legal experts, potentially streamlining workflow and alleviating the global backlog of pending cases. In response, we have developed a novel computational model that reliably generates calibrated probability estimates within the legal domain. Experiments on two public datasets showed that BayesJudge correctly identifies instances of uncertain predictions, significantly enhancing both effectiveness and efficiency. This streamlined process demands less effort, allowing focused attention exclusively on predictions where the model indicates uncertainty. In addition, our proposed optimal solution achieved a remarkable 20\% enhancement in accuracy when incorporating advanced legal text, and a notable 27\% improvement with the application of the extended legal text technique. Besides, there was a significant decrease in the Brier score, underscoring an augmented level of confidence in the model's predictions compared to its prior state.

While we have showcased the significant advantages of BayesJudge, it's crucial to acknowledge its inherent limitations. As data volume increases, BayesJudge tends to be overconfident, even in instances of erroneous predictions. This overconfidence arises despite the general desirability of ample data for robust machine learning model training. To address this, we can leverage hyper-prior optimization, effectively tuning the model's priors to mitigate overconfidence. Additionally, incorporating proven text processing techniques, commonly employed in other Bayesian models like topic models, can further enhance accuracy. Notably, our results demonstrate BayesJudge's optimal performance in resource-constrained scenarios, frequently encountered in the legal domain.

BayesJudge leverages the synergy of Bayesian inference and kernel methods, demonstrably enhancing its confidence in predictions. We have achieved notable improvements in the Brier score, a key metric for probabilistic forecast accuracy. These promising results pave the way for exploring applications in other domains with similar complex data, such as healthcare. BayesJudge performs remarkably with its natural adaptability. By fine-tuning the priors, it can seamlessly transfer its expertise to new domains, minimizing the need for costly adjustments. Simply crafting a dedicated posterior inference engine for the priors unlocks remarkable domain-agnostic capabilities.

\bibliographystyle{IEEEtran}
\bibliography{biblio}

\end{document}